%% file: main.tex
\documentclass[11pt]{article}

\usepackage[margin=1in]{geometry}
\usepackage[T1]{fontenc}
\usepackage[utf8]{inputenc}
\usepackage{microtype}
\usepackage{booktabs}
\usepackage{graphicx}
\usepackage{amsmath}
\usepackage[round]{natbib}
\usepackage[colorlinks=true,linkcolor=blue,citecolor=blue,urlcolor=blue]{hyperref}
\usepackage{xspace}
\usepackage{pifont}
\newcommand{\cmark}{\ding{51}}
\newcommand{\xmark}{\ding{55}}

\newcommand{\veracium}{Veracium\xspace}

\title{Ground Truth First: A Longitudinal Evaluation Instrument for Agent
Memory, and the Tenure Crossover in Memory-Architecture Rankings}

\author{Quentin Spencer\\
Independent Researcher\\
\texttt{ORCID 0009-0004-7318-2650}}

\date{}

\begin{document}
\maketitle

\begin{abstract}
Benchmarks for LLM-agent memory typically generate conversations first and
extract answer keys from them afterwards --- a pipeline with documented
label-error and contamination problems --- and they overwhelmingly measure
short interaction histories. We invert that pipeline: a deterministic, seeded
life-script sampler emits facts --- with validity intervals, volatility
classes, and source channels --- before any text exists; an LLM renderer
writes chat and email from per-event fact manifests; a fidelity verifier
confirms every planted fact; and questions are instantiated mechanically from
the script, so gold answers are script-valid by construction and separately
validated for answerability in the rendered corpus. The fully synthetic,
fictionalized corpus ($\sim$380 validated questions, 15 question types)
embeds features absent from the benchmarks we survey: per-fact validity
intervals, sent/received trust distinctions, injection probes inside a benign
harness, and as-of-date question sets. Benchmarking five memory architectures
against a no-memory control under a controlled protocol (fixed answerer,
versioned LLM judge, three stochastic replicates, two horizons), we find backend rankings
invert with history length: the budgeted curated-map memory that leads at
three weeks of history loses recall of evicted content by nine weeks
(96\%$\rightarrow$72\%), while a provenance-typed graph rises to 90\% ---
short-horizon benchmarks can mis-rank the very systems a long-lived assistant
needs. A full-rendered-history baseline sharpens the point: in this corpus
it tied or exceeded the tested memory systems at the short horizon ---
small-history evaluations may fail to distinguish memory architecture from
simply reading everything --- while at nine weeks no judge-independent
accuracy winner emerged between the layered system and reading the entire
raw transcript, at roughly half the read cost for the layered system. The
crossover itself is judge-robust: under complete re-judging by a
cross-family judge, the ranking inversion is positive for all six users
(exact $p=0.031$); not every pairwise margin survives the same test. Two further results: write-stage quality is a strong correlate of
downstream quality (questions backed by weakly written facts failed 24\%
vs.\ 2\% on clean writes); and injection resistance tracked the
preservation of provenance boundaries: the provenance-typed graph produced
zero unsupported assertions across all planted, non-adaptive injection
probes (14 unique questions, 42 answer trials) --- as did a raw transcript
that preserves source framing --- while flattened assertional memory
stores failed on a subset. A layered architecture --- unbounded
provenance-preserving store, LLM-curated budgeted view, focused
per-question retrieval --- performed best among the memory architectures in
both regimes (96.8\% short-horizon) and is released as \veracium, an
open-source library
(\texttt{pip install veracium}), together with the corpus generator and
evaluation harness.
\end{abstract}

\input{01_intro}
\input{02_related}
\input{03_method}
\input{05_results}
\input{06_back}

\bibliographystyle{plainnat}
\bibliography{refs}

\input{07_appendix}

\end{document}

%% file: 01_intro.tex
\section{Introduction}
\label{sec:intro}

Agents are becoming long-lived. An assistant that drafts email, triages a
mailbox, or manages a project accumulates months of interaction with one user
--- and what it should \emph{remember} becomes as consequential as what it
can reason about. The engineering response has been a wave of memory systems:
extraction pipelines over vector stores \citep{chhikara2025mem0}, temporal
knowledge graphs \citep{rasmussen2025zep}, self-editing agent memory
\citep{packer2023memgpt}, budgeted context maps \citep{gu2026peek}. The
evaluation response has not kept pace. The standard benchmarks generate
conversations first and extract answer keys afterwards --- a pipeline with
audited label errors \citep[6.4\% of LoCoMo's answer
key;][]{penfield2026locomoaudit,northcutt2021labelerrors} --- use real-world
entities that leak to pretraining (frontier models answer 70--85\% from
priors on MemGym's diagnostic; \citealp{xu2026memgym}), saturate at the top
of leaderboards \citep{he2026memoryarena}, and measure almost exclusively
\emph{short} interaction histories, the regime where full-context answering
is still competitive \citep{convomem2025}.

This paper asks what an agent-memory evaluation looks like when ground truth
comes first, and what it reveals when it runs long. We built a fully
synthetic, fictionalized corpus in which a deterministic life-script sampler
emits facts --- with validity intervals, volatility classes, and
channel/authorship provenance --- \emph{before any text exists}; an LLM
renderer writes chat and email from per-event fact manifests; a fidelity
verifier confirms every planted fact; and questions are instantiated
mechanically from the script, so gold answers are script-valid by
construction and separately audited for rendered-corpus answerability.
On this instrument we benchmark five memory architectures against a
no-memory control under a fixed answerer, a versioned LLM judge treated as
part of the benchmark, and threefold stochastic replication, across two history
horizons.

Three findings organize the results.

\paragraph{Rankings invert with history length.} The budgeted curated-map
memory that wins the short horizon (94.2\%) loses recall of evicted early
content as history grows (96.3\%$\rightarrow$72.2\% on early-chapter
probes), while a provenance-typed graph rises to 90.4\% and a layered hybrid
to 93.2\% by nine weeks. The aggregate ranking crossed over, all six users
show graph $\geq$ map by week~9, and the interaction is judge-robust:
under the primary judge, five of six per-user interaction effects are
positive (exact six-user randomization $p=0.063$, resolution-limited);
under a complete re-judging by a cross-family judge, all six are positive
(exact $p=0.031$, the test's minimum; \S\ref{sec:crossover}). Short-horizon benchmarks
--- the field's norm \citep{mem0report2026} --- can mis-rank the very
systems a long-lived assistant needs, and recent disputes about whether the
write path matters at all \citep{derehag2026smartsearch,yuan2026bottlenecks}
look different once horizon is treated as a variable. Two history baselines
locate the effect precisely: full rendered history tied or exceeded the
tested memory systems on the short horizon (confirming the sub-threshold
regime of \citealp{convomem2025} directly), while at nine weeks no
judge-independent accuracy winner emerged --- the layered system leads
under the primary judge, trails under the cross-family judge --- at roughly
half the per-question read cost, with read cost growing far more slowly
than full history's; a recency window matched to the memory system's mean
short-horizon read budget collapses in both regimes. What memory buys at
the same mean read budget is selectively constructed, structured context;
what it buys against full history at tenure is drastically lower
input-token growth, with no judge-independent accuracy winner at week nine
(\S\ref{sec:cost}).

\paragraph{In these systems, write-stage quality was a strong correlate of
downstream quality.} A write-stage audit against script ground truth shows the
two-stage curated writer produces ${\sim}2\times$ cleaner memories than
single-call extraction (half the audited errors: 4.0 vs.\ 8.5 per audit; fact recall
84.5\% vs.\ 74.6\%), write errors align
with downstream QA failures fact-for-fact, and a layered design ---
unbounded provenance-preserving store below, LLM-curated budgeted view
above, focused retrieval as supplement --- performed best among the memory
architectures on both horizons (96.8\% short-horizon, with no significant
primary-judge difference from full rendered history there; the only system
top-tier at every tenure). These are complete engineered systems that differ in write path,
representation, and token spend together, so we claim a strong association,
not a causal isolation (\S\ref{sec:writequality}).

\paragraph{Observed injection resistance tracked provenance boundaries ---
and accuracy does not confer caution.} The provenance-typed graph --- third-party claims stored
as claims, with the claimant as source --- blocked every planted,
non-adaptive injection probe (14 unique questions over 42 answer trials)
and every rung of a plausibility-graded
attack ladder with no added defense, while flattened assertional stores
breached on mundane attacks --- and a raw transcript that preserves source
framing also resisted (\S\ref{sec:cost}): the operative boundary is
provenance surviving representation, not graph topology per se
(\S\ref{sec:security}). And
although the accuracy winner also confabulates least in absolute terms, it
almost never abstains: 93.9\% of its errors are confidently asserted wrong
specifics under a deterministic answer-text classifier (8 abstentions in
972 answers), and the one surviving injection
leak re-enters through the episodic channel --- together specifying,
precisely, the evidence-grounded abstention gate a production system needs
(\S\ref{sec:security}).

\paragraph{Contributions.}
\begin{enumerate}
\item A ground-truth-first evaluation instrument for longitudinal,
  multi-channel, adversarial agent memory --- validity intervals, volatility
  classes, sent/received trust, injection probes in a benign harness,
  as-of-date question sets --- a combination absent from the published
  benchmarks we surveyed as of July 2026 (\S\ref{sec:related},
  Table~\ref{tab:gap}).
\item A controlled comparison of five memory architectures plus control,
  across two horizons and three stochastic replicates, under a fully
  logged protocol (52{,}797 LLM calls including validation).
\item The tenure crossover, with its mechanism (eviction of early-epoch
  content) isolated and replicated.
\item The write-quality result, with a fact-level audit linking write errors
  to downstream failures.
\item The observed pattern that injection resistance tracked whether
  provenance boundaries survive representation, plus the
  confident-confabulation failure mode that motivates read-time
  abstention.
\item \veracium, an open-source, provenance-aware memory library that
  operationalizes the winning design, and the corpus generator and harness,
  released for reproduction and extension.
\end{enumerate}

%% file: 02_related.tex
\section{Related work}
\label{sec:related}

\subsection{Agent memory architectures}

Long-term memory for LLM agents has three architectural lineages.
\emph{Curated working views}: MemGPT pages a small working context over a
larger store \citep{packer2023memgpt}; Generative Agents synthesize episodes
into reflections \citep{park2023generative}; PEEK maintains a fixed-budget
``context map'' via a Distiller, Cartographer, and priority Evictor
\citep{gu2026peek} --- the design our curated-map backend instantiates.
\emph{Retrieval over unbounded stores}: vector retrieval with decay
\citep{zhong2023memorybank}, graph-structured memory with associative
retrieval \citep{gutierrez2024hipporag,xu2025amem}, and commercial systems
--- Mem0's extraction pipeline \citep{chhikara2025mem0} and Zep's bi-temporal
knowledge graph \citep{rasmussen2025zep}. \emph{Layered combinations}:
TriMem keeps raw segments, atomic facts, and synthesized profiles at three
granularities \citep{trimem2026}; TierMem composes an immutable raw-log
store with a summary tier and a sufficiency router that escalates between
them \citep{zhu2026tiermem}; \citet{roynard2026knowledgelayer} argues from
cognitive-architecture principles that different knowledge layers need
different persistence semantics. Our layered result (\S\ref{sec:layered}) is
consistent with this direction but differs from its nearest neighbor TierMem
in three measurable ways: a \emph{budgeted, LLM-curated} working view (not
summaries), a security evaluation, and a tenure analysis --- the latter two
being where the layering earns its keep. Industry systems have converged on
adjacent shapes --- Perplexity's ``Brain'' compiles a context-graph work
memory into an overnight-synthesized wiki \citep{perplexity2026brain} ---
which we cite as evidence of the pattern's practical currency, not as
evaluated results.

\subsection{Benchmarks, and the case for ground-truth-first generation}
\label{sec:related-benchmarks}

The field's standard memory benchmarks are conversation-first: LoCoMo
extracts QA keys from generated long-term dialogues
\citep{maharana2024locomo}; LongMemEval curates questions into scalable chat
histories \citep{wu2024longmemeval}; MemoryAgentBench evaluates incremental
multi-turn competencies \citep{hu2026memoryagentbench}; industry surveys of
practice treat this set --- with BEAM \citep{tavakoli2025beam} --- as the
de-facto standard \citep{mem0report2026}. Their weaknesses are documented.
LoCoMo's answer key fails an independent audit at 6.4\% (99/1{,}540
score-corrupting errors, with an LLM judge accepting 62.81\% of deliberately
wrong answers) \citep{penfield2026locomoaudit} --- a non-peer-reviewed audit
with public, reproducible artifacts, corroborated by an open issue on the
benchmark's own repository; label-error rates of this order are consistent
with peer-reviewed findings across ML test sets
\citep{northcutt2021labelerrors}. Real-entity content leaks to pretraining:
MemGym shows frontier models score 0.70--0.85 on its diagnostic without
fictionalization \citep{xu2026memgym}, part of a broader contamination
picture \citep{xu2024contamination}. ConvoMem shows that below ${\sim}150$
conversations full-context answering beats retrieval memory
\citep{convomem2025} --- so short-horizon benchmarks also compress the very
differences they aim to measure. Newer benchmarks extend \emph{what} is
measured --- proactive memory-to-action \citep{shen2026mem2actbench},
memory-organization structure \citep{shutova2026structmemeval}, and
interdependent multi-session tasks, where MemoryArena finds that agents
near-saturated on LoCoMo fail badly once memory must guide sequential
decisions \citep{he2026memoryarena} --- without changing the
conversation-first pipeline.

Ground-truth-first generation itself is not new: RIKER builds RAG corpora
truth-first for static document QA \citep{roig2026riker}, and OrgForge-IT
applies the same philosophy to security telemetry \citep{flynt2026orgforge};
BEAM generates million-token dialogue histories with validated questions
\citep{tavakoli2025beam}. Our contribution is extending the inversion to
\emph{longitudinal, multi-channel, adversarial agent memory}: per-fact
validity intervals, volatility classes, a sent/received trust distinction,
injection probes inside a benign harness, and as-of-date question sets for
tenure curves. Table~\ref{tab:gap} locates each feature against the
published benchmarks; we did not find any of the five features in the
surveyed benchmarks and released artifacts as of July 2026.
(MemStrata implements temporal validity \emph{in a system}
\citep{yadav2026memstrata}, which sharpens the point: validity intervals
exist in products but not yet in any benchmark's ground truth.)

\begin{table}[t]
\centering
\small
\caption{Benchmark feature gap analysis. GTF = ground-truth-first
generation; VI = per-fact validity intervals; VC = volatility classes; TR =
sent/received trust distinction; IP = injection probes in a benign harness;
AO = as-of-date question sets; FE = fictionalized entities; LH =
longitudinal (multi-week/multi-session) horizon. Judgments verified against
abstracts and released artifacts (July 2026); $\pm$ = partial or unclear.}
\label{tab:gap}
\begin{tabular}{lcccccccc}
\toprule
Benchmark & GTF & VI & VC & TR & IP & AO & FE & LH \\
\midrule
LoCoMo \citep{maharana2024locomo}            & \xmark & \xmark & \xmark & \xmark & \xmark & \xmark & \cmark & \cmark \\
LongMemEval \citep{wu2024longmemeval}        & \xmark & \xmark & \xmark & \xmark & \xmark & \xmark & \xmark & \cmark \\
MemoryAgentBench \citep{hu2026memoryagentbench} & \xmark & \xmark & \xmark & \xmark & \xmark & \xmark & $\pm$ & $\pm$ \\
ConvoMem \citep{convomem2025}                & \xmark & \xmark & \xmark & \xmark & \xmark & \xmark & \cmark & \cmark \\
MemGym \citep{xu2026memgym}                  & \xmark & \xmark & \xmark & \xmark & \xmark & \xmark & \cmark & \cmark \\
Mem2ActBench \citep{shen2026mem2actbench}    & \xmark & \xmark & \xmark & \xmark & \xmark & \xmark & $\pm$ & \cmark \\
MemoryArena \citep{he2026memoryarena}        & \xmark & \xmark & \xmark & \xmark & \xmark & \xmark & $\pm$ & \cmark \\
RIKER \citep{roig2026riker}                  & \cmark & \xmark & \xmark & \xmark & \xmark & \xmark & \cmark & \xmark \\
BEAM \citep{tavakoli2025beam}                & \xmark & \xmark & \xmark & \xmark & \xmark & \xmark & $\pm$ & \cmark \\
\midrule
\textbf{Ours}                                & \cmark & \cmark & \cmark & \cmark & \cmark & \cmark & \cmark & \cmark \\
\bottomrule
\end{tabular}
\end{table}

\subsection{Does the write path matter? A horizon-dependent picture}
\label{sec:related-writepath}

Two recent studies find that retrieval and ranking dominate end-to-end
memory quality while write-time strategy barely matters: SmartSearch reports
that a deterministic ranking pipeline beats LLM-structured stores
\citep{derehag2026smartsearch}, and a diagnosis study finds accuracy spans
${\sim}20$ points across retrieval methods but only 3--8 across write
strategies \citep{yuan2026bottlenecks}. Both measure on short-horizon
benchmarks (LoCoMo, LongMemEval-S). Our results agree \emph{in that regime}
--- at ${\sim}30$--60 records per user, write-path differences are
compressed (\S\ref{sec:bakeoff}, cf.\ ConvoMem's threshold) --- and disagree
sharply outside it: the write-quality audit aligns with downstream failures
fact-for-fact (\S\ref{sec:writequality}), and the tenure crossover
(\S\ref{sec:crossover}) shows the regime where curation and time-aware
structure dominate is precisely the regime short-horizon benchmarks never
enter. The two findings are consistent once horizon is treated as a
variable --- reconciling them descriptively is one of this paper's
contributions; a causal resolution needs the crossed ablation named in
\S\ref{sec:writequality}. Two methodological
neighbors sharpen the picture: MemDelta shows memory-method rankings flip
with the answerer's model family and even the embedding model
\citep{wang2026memdelta} --- confounds our protocol excludes by holding the
answerer fixed and making only differential claims (the residual
single-answerer scope is discussed in \S\ref{sec:threats}) --- and MemTrace
traces operation-level error propagation through memory evolution
\citep{deng2026memtrace}, which our fact-for-fact write audit against
scripted ground truth complements from the other end.

Relatedly, Anatomy of Agentic Memory shows memory-system rankings are
fragile to backbone and benchmark choice \citep{jiang2026anatomy}, and a
systematic study concludes no single memory architecture dominates across
workloads \citep{zhou2026agentnative}. We strengthen this line with a
\emph{specific, mechanistic} instance: a measured ranking inversion in a
single variable (history length), with its cause isolated (eviction of
early-epoch content, \S\ref{sec:crossover}). To our knowledge the
history-length inversion itself is not shown in prior work.

\subsection{Memory security: poisoning, provenance, and where defenses
live}
\label{sec:related-security}

Indirect prompt injection via retrieved content is the founding attack class
\citep{greshake2023injection}; AgentPoison demonstrates poisoning of the
memory/RAG store itself, with retrieval attack success of 82\% and
end-to-end target impact of 62.6\% \citep{chen2024agentpoison}, and
standards bodies treat the risk as unresolved by RAG design
\citep{owasp2025llm01}. A 2026 wave sharpens the memory-specific threat:
MPBench benchmarks memory-poisoning attacks systematically
\citep{dash2026mpbench}; sleeper attacks plant content that stays dormant
across sessions before activation \citep{pulipaka2026sleeper,li2026plant};
TMA-NM proves content- and lineage-based defenses unsound under laundering
(via the agent's own summarization, trusted-tool echo, or manufactured
corroboration) and achieves 0\% attack success with non-malleable,
origin-bound authority --- an engineered enforcement mechanism with
machine-checked guarantees \citep{louck2026tmanm}. A lifecycle survey of
long-term-memory security concludes that defenses ``cannot be retrofitted at
retrieval or execution time alone'' but must be anchored in storage-time
provenance, versioning, and policy-aware retention \citep{lin2026ltmsecurity};
conceptual governance frameworks propose consistency verification and access
control before consolidation \citep{lam2026ssgm}; post-hoc auditing detects
poisoned memories after the fact \citep{tan2026memaudit} (distinct from
\textsc{MemAudit} the write-quality evaluation protocol,
\citealp{bhargava2026memaudit}). Provenance \emph{pointers} also appear for
non-security ends (narrative reconstruction in SEEM, \citealp{lu2026seem}).

Our security result (\S\ref{sec:security}) is complementary and deliberately
narrower than TMA-NM's: we show that an \emph{ordinary provenance-typed
store} --- no authority mechanism, no added enforcement --- already provides
structural quarantine (zero assertions across 14 unique probes over 42
trials; 0/6 on a plausibility-graded ladder),
because third-party claims are represented as claims rather than facts. This
is the ``storage-time provenance'' the survey calls for, observed as a free
dividend of representation. We are explicit about scope: our probes are
planted, not adaptive; relation-level attacks on graph memories remain open;
and the laundering channels TMA-NM formalizes are real --- our own episodic
leak (\S\ref{sec:security}) is one of them.

\subsection{Supersession, temporal knowledge, and knowledge editing}

Fact updating has two lineages: knowledge editing in weights
\citep[ROME/MEMIT;][]{meng2022rome,meng2023memit}, evaluated by MQuAKE
including its temporal variant \citep{zhong2023mquake}, and temporal
invalidation in stores \citep[Zep/Graphiti's bi-temporal
edges;][]{rasmussen2025zep}, deterministic freshness recipes
\citep{reddy2026freshness}, and MemStrata's supersession rule
\citep{yadav2026memstrata}. Update handling is where deployed systems
measure worst --- MemoryAgentBench's FactConsolidation single-hop: Mem0
18.0, Zep/Graphiti 7.0 \citep[v4, Table~3;][]{hu2026memoryagentbench},
independently corroborated by \citet{reddy2026freshness} --- which motivates
our supersession design and its evaluation (\S\ref{sec:representation}). A
concurrent diagnosis confirms the gap and its length-dependence: bounded
self-maintained memory trails full context by 15 points on LongMemEval's
update subset, degrading from 68\% to 28\% as conversations grow $24\times$
\citep{patel2026supersede} --- a single-system degradation curve consistent
with, but distinct from, our \emph{cross-architecture ranking inversion}
(\S\ref{sec:crossover}). Our contribution to this line is a read-time
result: explicit bi-temporal versioned edges \emph{underperform}
prose-inline history at answer time (72/81 vs.\ 79/81), because rendering
superseded edges returns stale values to the answerer's view.

\subsection{Confabulation, abstention, and judged evaluation}

Abstention when evidence is absent is a recognized gap: semantic-entropy
methods detect confabulations \citep{farquhar2024semantic}, models carry
usable self-knowledge signal \citep{kadavath2022know}, and surveys catalogue
abstention methods \citep{wen2024abstention}; LongMemEval includes an
abstention ability \citep{wu2024longmemeval}. We add a measurement that
sharpens the need: under a deterministic answer-text classifier, our most
\emph{accurate} system almost never abstains when wrong --- 93.9\% of its
judged errors classify as wrong-specific rather than abstention, vs.\
62.7\% for the least accurate --- even though it confabulates least in
absolute terms under the primary judge; accuracy does not confer
calibrated abstention (\S\ref{sec:security}). For judged
evaluation, we follow the LLM-judge validity literature
\citep{zheng2023judging,wang2024notfair}: the judge is versioned as part of
the benchmark, kept in a separate model family from the corpus generator,
and all headline claims are differential (rankings, crossovers) rather than
absolute.

%% file: 03_method.tex
\section{The instrument}
\label{sec:instrument}

\subsection{Design principle: ground truth first}

Benchmarks for conversational and agent memory are typically built
conversation-first: dialogues are generated (or collected), and answer keys
are then extracted from them. This pipeline has two documented failure
modes. First, extracted keys are wrong at material rates --- an independent
audit of LoCoMo \citep{maharana2024locomo} found 6.4\% of its answer key
score-corrupting (99/1{,}540 questions), with an LLM judge that accepted up
to 63\% of deliberately wrong answers \citep{penfield2026locomoaudit};
label-error rates of this order are consistent with peer-reviewed audits
across ML test sets \citep[avg.\ 3.3\%;][]{northcutt2021labelerrors}.
Second, when benchmark entities are real, frontier models answer substantial
fractions of ``memory'' questions from pretraining alone --- MemGym measures
70--85\% on its diagnostic without fictionalization \citep{xu2026memgym} ---
so a memory score partly measures the model's priors, not its memory
\citep{xu2024contamination}.

Our instrument inverts the pipeline: \textbf{ground truth exists before any
text.} A deterministic, seeded life-script sampler emits a timeline of facts
for a fictional user --- each fact carrying a \emph{validity interval} (when
it is true), a \emph{volatility class} (permanent $\rightarrow$ durable
$\rightarrow$ slow-changing $\rightarrow$ transient $\rightarrow$
ephemeral), and a \emph{channel} (stated in chat; or arriving by email, with
sent vs.\ received distinguished --- received mail is third-party-authored
content and therefore both a lower-trust evidence class and an injection
surface). An LLM renderer --- Amazon Nova Pro, deliberately a different
model family from the answerer and judge --- turns each scripted event into
a chat session or an email against a per-event \emph{fact manifest}. A
fidelity verifier then checks that every planted fact actually appears in
the rendered text (regenerate on miss; ship-with-warning after retries).
Questions are instantiated mechanically from the script, never from the
rendered text, so gold answers are \emph{script-valid by construction};
whether the rendered corpus actually supports each answer is separately
validated (\S 3.3). The fidelity verifier checks planted-fact
\emph{presence}, not renderer additions. The answerability audit is a
partial backstop --- it exposed one case of renderer-invented content in
practice --- but an open-book oracle can succeed despite unsupported
additions or accidental shortcuts, so it is not a comprehensive audit of
renderer precision; a two-sided renderer audit (planted-fact recall
\emph{and} unsupported-fact precision) remains future work
(\S\ref{sec:threats}).

Because the corpus is fully synthetic and fictionalized, it contains no real
personal data (an ethics asset), and entity-specific gold facts are
fictionalized and unavailable from public factual priors by design
\citep[a validity asset; cf.][]{xu2026memgym}.

\subsection{Corpus}

Two corpora share the generator:
\begin{itemize}
\item \textbf{Short horizon:} 14 users, 275 validated questions, dual
  chat/email channels, planted injection probes and staleness traps.
\item \textbf{Long horizon (tenure):} 6 users with nine-week histories, 108
  questions instantiated as \emph{as-of-date} sets at three checkpoints
  (weeks 3, 6, 9; 324 judged answers per backend per checkpoint), including
  early-chapter probes that specifically target the oldest material.
\end{itemize}

Total: 383 validated questions across 15 question types (three types add
\texttt{-early} variants in the tenure corpus): single-fact,
inference-only, episodic, temporal, multi-session, knowledge-update
(current and past), preference (current and application), expired-state,
conflict-resolution, work-memory, abstention, abstention-false-premise, and
injection-probe.

The corpus embeds features that we did not find in the memory benchmarks
and released artifacts we surveyed as of July 2026
(Table~\ref{tab:gap}): per-fact validity intervals; a
sent/received trust distinction on the email channel; injection probes
embedded in a benign harness rather than an adversarial one; transient
states with ground-truth end dates; and as-of-date question sets enabling
tenure curves.

\subsection{Answerability audit}

Following MemGym's protocol, a final validation stage gives an oracle the
\emph{complete} rendered corpus open-book and requires it to answer every
generated question; the standard judge scores the attempt, and failures are
excluded from the harness. 215/220 generated questions validated (97.7\%).
The five exclusions were not noise: they trace to a single named tension ---
a well-behaved oracle \emph{declines to assert} facts whose only evidence is
third-party-authored (a relative's email), which is precisely the behavior
that defeats injection. We return to this trust/recall trade-off in
\S\ref{sec:security}; the instrument surfaced it before any memory system
existed.

\subsection{Harness and judge}
\label{sec:harness}

All backend comparisons hold the answering model fixed
(\texttt{claude-haiku-4-5-20251001}) and include a no-memory ablation
control. The judge (\texttt{claude-sonnet-5}, versioned rubric) is treated
as \emph{part of the benchmark definition}; the renderer is
\texttt{amazon.nova-pro-v1:0} and the cross-family judge
\texttt{deepseek.v3.2} (both via Bedrock), with Titan-v2 embeddings for the
vector backend. All calls used provider-default decoding at run time
(July 2026; Bedrock region \texttt{us-east-1}; up to 2--3 retries on
transient failures and parse errors, recorded per call); every call is
logged (model, tag, token counts) and the logs are released. Provider
defaults are not archived by the providers, which we flag as a
reproducibility limitation. Rubric changes are versioned like code; the
judge is held fixed across backends so that shared error cancels in
differential claims \citep{zheng2023judging,wang2024notfair}; and the
corpus generator is a different model family from the answerer and judge.
Holding the judge fixed does not make \emph{all} its error common-mode ---
error that interacts with answer style is backend-correlated --- which
\S\ref{sec:threats} addresses. Every run is repeated as three independent
stochastic replicates (repeated API runs --- the providers expose no
random-seed control, so these are replicates, not seeded runs);
stochasticity in a single run masked entire defect classes in early
iterations. The corpus \emph{generator} is genuinely seeded and
deterministic; only the LLM calls are stochastic. Content-level
replication comes
from the users themselves: each user --- short-horizon and tenure alike ---
is an independent generator draw (persona, preference chain, fact values),
with tenure users of the same archetype sharing only a chapter-index
work-fact skeleton (\S\ref{sec:threats}). Per-cell model configurations are
recorded and released.

The original study logged 34{,}076 LLM calls end to end, with judging
consuming more calls than answering (15.0k vs.\ 12.7k) --- half the
instrument's spend is measurement, which we report as the honest price of
a judged benchmark. Review-driven validation added a further 18{,}721
calls (the history baselines and the complete cross-family re-judging,
including the vector follow-up), for 52{,}797 in the released logs.

\section{Study design}
\label{sec:design}

We benchmark five memory architectures plus the no-memory control under the
\S\ref{sec:harness} protocol. The bake-off backends consume one shared
distillation pass, but the complete systems differ together in write
procedure, representation, retrieval, and token spend
(\S\ref{sec:writequality}, \S\ref{sec:cost}): the experiment compares
\emph{complete memory-system implementations} under a shared corpus,
answerer, and evaluation protocol --- not isolated components.

\begin{enumerate}
\item \textbf{Curated map} --- a budgeted Markdown context map maintained by
  an LLM curator \citep[the PEEK shape;][]{gu2026peek}: distill
  $\rightarrow$ cartograph $\rightarrow$ evict under budget.
\item \textbf{Vector} --- Titan-v2 embeddings, top-$K$ retrieval over
  unbounded records.
\item \textbf{Graph} --- a typed labeled-property graph; entities,
  relations, and provenance-typed edges; entity-matched subgraph retrieval.
  Third-party claims exist only as \texttt{third\_party\_claim} edges with
  the claimant as source.
\item \textbf{Hybrid v1} --- a deterministic graph$\rightarrow$map
  compilation (negative result, \S\ref{sec:hybrid1}).
\item \textbf{Hybrid v2 (layered)} --- the graph's typed edges plus the
  distiller's episode records (each store's cleaner half), compiled once per
  user by an LLM cartographer into a budgeted wiki, supplemented by
  per-question subgraph retrieval.
\end{enumerate}

Questions are scored by the versioned judge; we report totals, per-category
splits, and --- for the tenure corpus --- per-checkpoint accuracy on
as-of-date question sets.

%% file: 05_results.tex
\section{Results}
\label{sec:results}

\subsection{Short horizon: specialization, not a winner}
\label{sec:bakeoff}

The bake-off runs 275 questions (spanning 14 users) $\times$ 3 replicates $=$
825 judged answers per backend (Table~\ref{tab:bakeoff}).

\begin{table}[t]
\centering
\small
\caption{Short-horizon bake-off (825 judged answers per backend, 3 replicates
pooled). Vector's single category win (conflict-resolution, 42/42 vs.\
41/42 for both map and graph) is one answer inside replicate noise.}
\label{tab:bakeoff}
\begin{tabular}{lrrl}
\toprule
Backend & Total & \% & Signature strengths \\
\midrule
none (control) & 125/825 & 15.2 & --- \\
curated map    & 777/825 & 94.2 & multi-session 24/27, current-state \\
vector         & 751/825 & 91.0 & no signature strength \\
graph          & 769/825 & 93.2 & single-fact 119/120, work-memory 124/126, injection 42/42 \\
hybrid v1      & 738/825 & 89.5 & (negative result, \S\ref{sec:hybrid1}) \\
hybrid v2      & \textbf{799/825} & \textbf{96.8} & first by 22; single-fact 120/120, work-memory 126/126 \\
\bottomrule
\end{tabular}
\end{table}

The three recall targets of \S\ref{sec:intro} map onto the winners:
user-model facts and work knowledge are graph-shaped; current-state
synthesis is curated-map-shaped.

\subsection{Naive composition fails}
\label{sec:hybrid1}

Hybrid v1 --- a deterministic compile of the graph into a map --- scored
738/825 (89.5\%), \emph{below both parents}. The graph stored no narrative
for the compiler to use, and deterministic string-joins amplified extraction
noise that each parent had been quietly curating away (entity focus in the
graph; an LLM cartographer in the map). Composition of two curated stores is
not itself curation.

\subsection{The layered architecture leads the memory systems at the short
horizon}
\label{sec:layered}

Hybrid v2 scored 799/825 (96.8\%) --- first among the memory architectures
by 22 answers, with no significant primary-judge difference from the
full-rendered-history baseline (\S\ref{sec:cost}): single-fact
120/120, work-memory 126/126, conflict-resolution 42/42 (the curator
repaired the noisy chains that sank v1), and best-in-class inference-only
(76/81) and episodic (8/9). The lead is significant under paired,
cluster-aware analysis: vs.\ the map, 45/23 discordant items (exact McNemar
$p=0.010$; user-cluster bootstrap 95\% CI for the difference $[+0.2,+5.5]$
pp); vs.\ the graph, 37/7 ($p<10^{-4}$; CI $[+1.6,+6.1]$ pp). Under a
\emph{complete} cross-family re-judging (\S\ref{sec:threats}) the graph
margin holds (74/47, $p=0.018$) while the map margin keeps its direction
but loses significance (88/69, $p=0.15$) --- consistent with the narrow
cluster interval above; we therefore treat the hybrid-v2--map short-horizon
lead as a primary-judge result. Honest
remainders: multi-session integration 19/27 (behind the pure map's 24/27),
preference-current 34/42, and the wiki reintroduces a flat-text injection
surface (40/42 vs.\ the raw graph's 42/42; \S\ref{sec:security}). Because
hybrid v2 was designed after observing v1's failure on this corpus, we treat
this number as benchmark-informed (\S\ref{sec:threats}); the long-horizon
result below is the stronger evidence.

\subsection{The tenure crossover}
\label{sec:crossover}

\begin{figure}[t]
\centering
\includegraphics{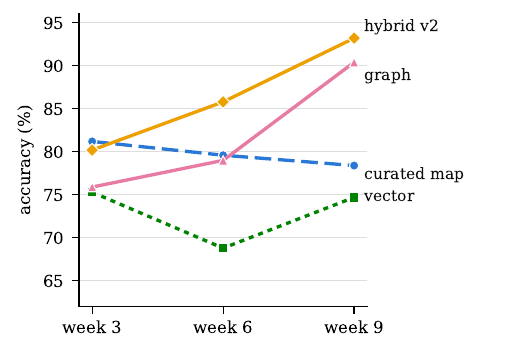}
\caption{The tenure crossover: aggregate accuracy by history length
(6 users $\times$ 3 replicates, $n{=}324$/cell; the no-memory control,
16.0--16.7\%, is omitted for scale). The curated map leads at week~3; the
graph and layered hybrid invert the ranking by week~9.}
\label{fig:crossover}
\end{figure}

The tenure corpus runs 6 users $\times$ 3 replicates ($n{=}324$ per cell) with
as-of-date question sets at weeks 3, 6, and 9
(Table~\ref{tab:crossover}, Figure~\ref{fig:crossover}).

\begin{table}[t]
\centering
\small
\caption{Tenure accuracy (\%) by checkpoint (6 users $\times$ 3 replicates,
$n{=}324$/cell).}
\label{tab:crossover}
\begin{tabular}{lrrr}
\toprule
Backend & week 3 & week 6 & week 9 \\
\midrule
curated map & \textbf{81.2} & 79.6 & 78.4 \\
vector      & 75.3 & 68.8 & 74.7 \\
graph       & 75.9 & 79.0 & \textbf{90.4} \\
hybrid v2   & 80.2 & 85.8 & \textbf{93.2} \\
none        & 16.0 & 16.4 & 16.7 \\
\bottomrule
\end{tabular}
\end{table}

\textbf{Backend rankings invert with history length.} The curated map leads
the graph by 17 answers at week 3; they effectively tie at week 6 (a
2-answer gap); the graph leads by 39 at week 9 --- and every one of the six
users individually shows graph $\geq$ map at week 9. The week-3 and week-9
differences are each individually reliable: week-3 map $>$ graph (32/15
discordant, exact McNemar $p=0.019$; user-cluster bootstrap 95\% CI
$[+1.2,+9.0]$ pp) and week-9 graph $>$ map (57/18, $p<10^{-4}$; CI
$[+6.5,+19.4]$ pp). Separate significance at each end, however, is not
itself evidence that the difference \emph{changed}; the direct test is the
architecture$\times$tenure interaction,
$(\text{graph}-\text{map})_{\text{wk9}} -
(\text{graph}-\text{map})_{\text{wk3}}$: $+17.3$ pp on average, positive
for five of six users and null for one (per-user effects $+37.0, +16.7,
+9.3, 0.0, +20.4, +20.4$; user-cluster bootstrap CI $[+8.3,+26.2]$ pp),
with an exact six-user sign-flip randomization $p=0.063$ --- suggestive but
resolution-limited: six clusters with one null user cannot reach
conventional significance on this test. Under the complete cross-family
re-judging (\S\ref{sec:threats}), however, the interaction
\emph{strengthens}: $+24.1$ pp on average, positive for all six users
(per-user $+40.7, +20.4, +16.7, +13.0, +27.8, +25.9$; exact sign-flip
$p=0.031$, the test's minimum), with both endpoint margins significant
($p=0.010$ week 3, $p<10^{-5}$ week 9). The crossover is judge-robust at
the cluster level. Hybrid v2 is the overall long-horizon leader
(86.4\%; CI over the map $[+3.6,+10.1]$ pp), the only backend top-tier at
every tenure, and its interaction over the map is $+15.7$ pp with all six
users positive (exact sign-flip $p=0.031$, the minimum achievable).

\begin{figure}[t]
\centering
\includegraphics{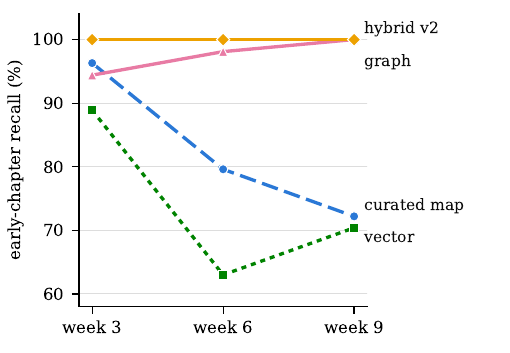}
\caption{Early-chapter (evicted-content) recall by history length: the
three \texttt{-early} question types pooled, $n{=}54$ per cell. The
budgeted map's eviction consumes the oldest material; the unbounded graph
and the layered hybrid protect it.}
\label{fig:earlychapter}
\end{figure}

The mechanism is recall of evicted content
(Figure~\ref{fig:earlychapter}). On the early-chapter probe (the three
\texttt{-early} question types pooled; $n{=}54$ per cell) the map falls
96.3\% $\rightarrow$ 72.2\% as budget eviction consumes the oldest material
(13/0 discordant items week~3$\rightarrow$9, $p=2\times10^{-4}$), while the
graph rises 94.4\% $\rightarrow$ 100\% and hybrid v2 holds 100\% at every
checkpoint. A full-rendered-history baseline (\S\ref{sec:cost}) holds the
early-chapter probe at 160/162 across all checkpoints --- when nothing is
evicted, old content survives --- confirming that the map's loss is caused
by eviction, not by history length itself. The map's \emph{aggregate}
decline is mild ($-2.8$ pts) because most question types are
tenure-insensitive --- which is exactly why short-horizon benchmarks miss
the effect. Vector's unbounded
pool also failed the probe (88.9\% $\rightarrow$ 70.4\%): similarity
retrieval buries old records under similar newer content. Unbounded storage
without time-aware retrieval is not an episodic memory.

Short-horizon benchmarks --- the field's norm
\citep{maharana2024locomo,wu2024longmemeval,hu2026memoryagentbench,mem0report2026},
and our own \S\ref{sec:bakeoff} --- can mis-rank the architectures that a
long-lived assistant needs.

\subsection{Write-stage quality: a strong correlate of memory quality}
\label{sec:writequality}

A write-stage audit against script ground truth (66 audits: the 11
script-generated users $\times$ 3 replicates $\times$ both write paths) scored
the two-stage
curated writer at 84.5\% fact recall / 4.0 errors per audit against the
single-call graph extractor's 74.6\% / 8.5 --- half the audited error
count, bought with ${\sim}4.6\times$ the write-path tokens
(\S\ref{sec:cost}). Each audit scores one stored memory against the
life-script's atomic fact checklist: per-fact recall status
(correct/distorted/absent), an error list typed as
hallucination/distortion/misattribution, injection representation, and
volatility-class match. Scoring is by an LLM evaluator (the judge model)
run separately from --- and blind to --- downstream QA outcomes;
``fact-for-fact alignment'' means the audit's weak facts and the QA miss
categories matched at the level of individual fact keys --- and we can
quantify it, because the question templates and the audit checklist
reference the same script fields, giving a mechanical question--fact
mapping (17 of 20 templates map; abstention and injection probes have no
stored-fact key). Treating each (user, replicate, write path, question) as an
observation ($n{=}1{,}092$): a question whose underlying fact was written
weakly (distorted or absent) missed \textbf{24.2\%} of the time, against
\textbf{1.6\%} when the write was clean --- risk ratio ${\sim}15$, odds
ratio 19.6 (user-cluster bootstrap 95\% CI $[10.6, 50.3]$ over the 11
audited users; permutation $p<10^{-4}$, where weak-write labels are
shuffled within each user$\times$replicate$\times$write-path block with QA
outcomes held fixed, preserving each block's weak-label count;
$10^4$ permutations), consistent across both write paths (OR 16.3
curated, 25.6 extractor). 79\% of all QA misses had a weak-written
underlying fact. The scope is the association's own: the 11
script-generated short-horizon users and the 17 mechanically mapped
templates --- it has not yet been replicated on the tenure corpus or the
hand-built users. This quantifies the association --- it still does not
isolate causation (\S\ref{sec:threats}); hard facts could drive both
sides, and one measurement caveat applies: the audit evaluator is the same
model as the QA judge, so shared evaluator tendencies could induce
correlated measurement error --- the association's magnitude should not be
treated as independently validated until the audit labels receive human or
cross-family checking. The audit rubric, all 66 audit records, and the association script
are released; auditor--human agreement has not been measured and is
planned alongside the judge audit (\S\ref{sec:threats}). Write-stage weak facts aligned with downstream
QA miss categories fact-for-fact: two linked measurements --- separately
executed, but sharing the evaluator model and the scripted fact keys ---
one diagnosis. We state this as a strong association, not a causal isolation:
the compared writers differ in procedure \emph{and} compute, and hard or
ambiguous facts could contribute to both write and answer errors; a crossed
writer$\times$store ablation under matched token budgets is the experiment
that would settle causality.
Even the best write path damages ${\sim}15\%$ of facts; answering models
paper over the noise until a probe surfaces it. The residual error classes
are specific: detail distortion (dates, reasons), misattribution, and slot
errors (a permanent fact filed in a functional slot;
\S\ref{sec:representation}).

\subsection{Representation: supersession, episodes, volatility}
\label{sec:representation}

\paragraph{Supersede, never erase.} Records are soft-invalidated with
supersession pointers; the replacing record carries the prior value as
inline prose. On knowledge-update questions --- the category commercial
systems measure worst (FactConsolidation single-hop: Mem0 18.0,
Zep/Graphiti 7.0; \citealp[v4, Table~3]{hu2026memoryagentbench};
independently corroborated by \citealp{reddy2026freshness}) --- the curated
map scored 120/123 and hybrid v2 121/123 (42/42 on current-value questions;
misses concentrate in past-value recall). A surprise with design
consequences: prose-inline history \emph{beat} explicit bi-temporal
versioned edges at read time (79/81 vs.\ 72/81 on knowledge-update-past)
--- rendering superseded edges puts stale values back into the answerer's
view, a failure mode the KG lineage \citep{rasmussen2025zep} does not
anticipate at read time.

\paragraph{Episodes must exist by construction.} No LLM writer produced
dated episodic entries unprompted; once episodes became deterministic
per-event records, history questions flipped from systematic failure to
near-perfect --- and every architecture that discarded raw episodes after
distilling lost the past.

\paragraph{Volatility is real but hard to classify.} 120/161 (75\%) of
anchor facts landed in their expected class, and the failure mode is
structural --- a \emph{permanent} allergy filed into a functional
\texttt{health\_state} slot was superseded by a passing cold. Lifecycle
machinery operating correctly on mis-slotted facts silently destroys
permanent knowledge.

\subsection{Security: what the representation buys}
\label{sec:security}

A four-experiment sweep probed where security properties live.

\paragraph{Injection resistance tracked the record representation.} A
plausibility-graded ladder of six received-email attacks (absurd lottery
$\rightarrow$ generic debt $\rightarrow$ subscription-shaped $\rightarrow$
invoice-context $\rightarrow$ contact-impersonation spear-phish
$\rightarrow$ memory-poisoning directive) probed the stores. The graph
asserted 0 of 6: structural quarantine --- third-party claims exist only as
claim-edges with the claimant as source --- held across the entire gradient
with no added defense. The flat store asserted 3 of 6, and not the
extremes: the breaches span the subscription, invoice, and spear-phish
rungs, while the absurd and the overtly manipulative both tripped other
handling --- injection success is non-monotonic in plausibility. Labels are
not a defense: an ``unconfirmed''-flagged claim is still asserted
downstream. In the full benchmark the pattern reproduces quantitatively
across 14 unique planted injection questions (one per user), each
evaluated over three stochastic answer/judge trials: the graph produced
zero unsupported assertions in all 42 trials; the layered hybrid's
flat-text wiki failed on 2 of the 14 unique probes (2/42 trials), the
curated map likewise on 2 of 14. This is the attack class of indirect prompt injection via
retrieved content \citep{greshake2023injection} and memory poisoning
\citep{chen2024agentpoison}; descriptively, the graph and the full
transcript had no observed failures on the 14 probes, whereas the two
flattened stores each failed on two --- an observed pattern, not a
statistically established architecture difference (an exact paired test
with only two discordant unique probes cannot reach conventional
significance). Two scoping notes. First, the compared systems differ in
retrieval and serialization as well as representation, so
``representation property'' is an interpretation of the observed pattern
--- content labels failing while structural typing held at every rung ---
rather than a component-isolated ablation, which would hold the other
components constant. Indeed, the full-history baseline sharpens the
interpretation: a raw transcript --- also flat text --- likewise blocked
all 14 unique probes across 42 trials (\S\ref{sec:cost}), because the
received-mail trust framing survives
verbatim in it. What failed is \emph{flattened assertional} storage that
strips authorship boundaries in the act of summarizing; the operative
property appears to be provenance boundaries surviving representation,
whether structurally (claim-typed edges) or textually (intact source
framing) --- not graph topology per se. Second, the uncertainty must be
computed at the unique-probe level, not the trial level: zero failures on
14 unique probes bounds the per-probe assertion rate only below
${\sim}19\%$ (95\% upper bound, treating probes as independent --- itself
optimistic, since probes share a template family); the supported claim is
resistance to this documented, non-adaptive attack class --- not security
in general (\S\ref{sec:threats}).

\paragraph{Content-type quarantine is a write-gate property.} Obligations
and payment demands quarantine by \emph{content}, independent of sender
trust --- a spoofed known-correspondent demanding a wire transfer was
blocked even under graded sender trust, while benign third-party recall (a
relative's venue change) survived. Defense in depth that sender-trust alone
cannot provide.

\paragraph{Confabulation control is a read-gate nobody has.} We report both
the unconditional and the conditional view, because they disagree.
Unconditionally, the accuracy winner also confabulates least: over the 972
tenure answers, hybrid v2 asserts a wrong specific on 12.8\% of all
questions (124/972, with 8 abstentions) against vector's 17.0\% (165/972,
with 98 abstentions). The design-relevant
behavior is conditional: when hybrid v2 \emph{is} wrong, it almost never
abstains --- 93.9\% of its errors (124/132) are confidently asserted wrong
specifics, vs.\ 62.7\% for vector (165/263), classified by the
deterministic answer-text rule documented in
Appendix~\ref{app:crossjudge} --- so the best system's failures are
precisely the silent, confident kind a user cannot detect.
Accuracy does not confer calibrated abstention. The conditional pattern
persists under the complete cross-family re-judging of all four memory
systems: hybrid v2's confident-error share remains highest (85.4\%) and
vector's lowest (62.5\%, vs.\ 62.7\% under the primary judge --- the
original contrast is judge-robust), with graph 78.0\% and map 69.3\%
between them; unconditionally, all four sit near 23--24\% wrong-specific
under the stricter judge (Appendix~\ref{app:crossjudge}) --- and the
planned human adjudication uses a three-way rubric (correct /
wrong-specific / abstention) for exactly this reason. And the one injection that survives
structural quarantine re-enters through the \emph{episodic} channel
(``received an auto-renewal notice at \$890''), which is mentionable by
design. Both failures motivate the same missing mechanism: an
evidence-grounded abstention gate at answer time
\citep{farquhar2024semantic,kadavath2022know,wen2024abstention} that
refuses to assert claims whose only support is third-party-authored
episodes.

\subsection{Baselines: no history, full history, and a token-matched
window}
\label{sec:cost}

Memory-ablated deltas are 79--82 points short-horizon (94.2--96.8\% vs.\
15.2\%) and 70 points long-horizon (86.4\% vs.\ 16.4\%) against the
no-history control. To test what those deltas actually establish, we added
two history baselines under the identical prompt, answerer, judge, and
3-replicate protocol: \textbf{full rendered history} (every event, chronological
--- it fits in context at every horizon here) and a \textbf{token-matched
recency window} (most recent whole events under hybrid v2's mean read
budget of ${\sim}$1,807 tokens).

\textbf{Short horizon: full context ties or beats the best memory system.}
Full history scores 97.9\% (808/825) --- statistically indistinguishable
from hybrid v2's 96.8\% under the primary judge (25/16 discordant,
$p=0.21$; user-cluster CI $[-1.0,+2.9]$ pp), ahead of the curated map
($[+1.1,+6.6]$ pp), and significantly ahead of hybrid v2 under the
cross-family judge (98/28, $p<10^{-5}$; \S\ref{sec:threats}). This
directly confirms, rather than merely cites, the sub-threshold regime
\citep{convomem2025}: below ${\sim}150$ conversations, reading everything
is competitive, and small-history evaluations like this one may fail to
separate the tested memory systems from a trivial strategy. The recency
window, by contrast, collapses to 73.0\% --- work-memory 19/126,
preference-current 9/42, knowledge-update-past 44/81 --- at the same mean short-horizon token
budget, \emph{selectively constructed and structured} memory context
substantially outperformed a recency-only window: hybrid v2's 1,807 tokens
beat the same 1,807 tokens of recency by ${\sim}24$ points (the systems
differ in more than selection --- representation, compression, and
supplementary retrieval also change). (The window is matched to hybrid v2's \emph{mean}
short-horizon spend; realized spend varies per question and grows with
tenure, so the tenure-window comparison is under-matched --- per-question
and per-checkpoint matching is future work.) (Full history also
produced zero unsupported assertions on all 14 unique injection probes
across 42 trials: the received-mail trust framing travels with the
transcript.)

\textbf{Long horizon: no judge-independent winner, at lower read cost.} Full
history reaches 82.4/85.8/88.9\% across the three checkpoints with
near-perfect early-chapter recall (160/162) --- confirming the crossover's
mechanism from the other side: when nothing is evicted, old content
survives. At week 9, whether the layered system \emph{beats} full history
is judge-sensitive, and we report it as such: under the primary judge,
hybrid v2 leads 93.2\% vs.\ 88.9\% (13/27 discordant, item-level $p=0.039$;
per-user effects $+3.7, -1.9, +7.4, +5.6, +7.4, +3.7$ pp, five of six
positive, sign-flip $p=0.063$); under the complete cross-family re-judging,
full history edges ahead at week 9 (272 vs.\ 260 of 324, $p=0.18$) and
leads the tenure aggregate under that judge (79.7\% vs.\ 72.4\%;
Appendix~\ref{app:crossjudge}), while under the primary judge the tenure
aggregate is effectively tied (85.7\% vs.\ 86.4\%). The judge-robust
statement is that \emph{no judge-independent accuracy winner emerged at
nine weeks, while the layered system reads roughly half the tokens per
question} --- with read cost growing far more slowly than full history's
(below); the graph likewise shows no significant difference from full
history under either judge. (Non-significance is not equivalence: we claim
the absence of a robust winner, not demonstrated parity.) The
recency window fails both regimes (42.9--66.4\%; early-chapter 17/54 at
week 9). Remaining baselines for future work: raw-chunk RAG with a
competitive reranker, and an oracle-retrieval ceiling.

\textbf{Cost.} Marginal read cost is small: 677 (vector) to 1{,}807
(hybrid v2) input tokens per question --- input-token charges of
${\approx}\$0.002$/question at July-2026 list pricing for the answerer
(input tokens only; output tokens, storage, retrieval infrastructure, and
amortized write cost are excluded, and per-call wall-clock times are in
the released logs rather than analyzed here); full history spends the
entire corpus per question (${\sim}2.8$--$5.2$K tokens here), growing
linearly with corpus size, with no judge-independent accuracy advantage to
show for it (above); hybrid v2's read cost also grows ($+48\%$ by week 9)
but from a base roughly half full history's at week 9. Two structural
results: the budgeted map's
read cost is flat with tenure (1{,}183 $\rightarrow$ 1{,}182 tok/q) while
the unbounded winners' grows ($+48$--$57\%$ by week 9) --- forgetting is
what keeps the map cheap, and the tenure crossover is the price of that
economy; and write-side quality is bought with tokens (the curated writer
spends ${\sim}4.6\times$ the extractor's tokens to halve the audited error
count) --- but write cost amortizes over every future read, so
the economics and the write-quality association
(\S\ref{sec:writequality}) point at the same place: the write path.

%% file: 06_back.tex
\section{From findings to artifact: \veracium}
\label{sec:veracium}

The winning layered design is released as \veracium (\texttt{pip install
veracium}; MIT), a provenance-aware memory library whose design decisions
trace one-to-one to measured findings. The store is a typed graph with
provenance-carrying edges (\S\ref{sec:representation}'s
supersede-never-erase, with prior values rendered as inline prose rather
than versioned-edge dumps); raw episodes are deterministic per-event
records; an LLM-curated, budget-aware wiki is compiled \emph{above} the
store (\S\ref{sec:layered}, \S\ref{sec:writequality}); and an assertability
gate implements the missing mechanism \S\ref{sec:security} specifies: facts
whose only support is third-party-authored evidence are never assertable ---
they surface, if at all, as explicitly fenced unverified claims --- and the
gate extends through mixed-provenance events (content \emph{derived from}
third-party sources is trust-capped at the weaker source, closing the
summarization-laundering channel that \citet{louck2026tmanm} formalizes ---
a channel independently discovered by \veracium's first production consumer
on a real mailbox, and fixed at the representation level).

A distilled acceptance eval derived from the study's harness ships as the
library's regression gate (small fictional scenarios; asserts zero injection
assertions and a correctness floor), the full harness is released alongside
the corpus generator (\S\hyperref[sec:conclusion]{Reproducibility}), and an
opt-in robustness tier replays real chat corpora through the write path,
programmatically asserting
quarantine, isolation, and provenance invariants. An internal robustness
smoke test replayed 200 conversations (371 user turns) sampled with a fixed
seed from a public chat corpus (LMSYS-Chat-1M; license-gated, not
redistributed --- the shipped tier uses synthetic fixtures), with synthetic
injection content inserted into 15\% of turns and per-user isolation
probed across all users; the programmatic checkers reported zero
cross-user leaks, zero injection leaks, and zero malformed edges over all
turns and probe answers (and, in earlier iterations, caught two real crash
bugs --- the eval-driven loop working as intended). This is an internal
smoke test with scripted, non-blind detection --- not a product evaluation
or evidence of production-grade robustness; \veracium is the paper's
artifact, not its subject.

\section{Threats to validity}
\label{sec:threats}

\paragraph{Synthetic corpus.} Deliberate: real-entity benchmarks leak to
pretraining \citep{xu2026memgym,xu2024contamination}, and truth-first
generation is what makes gold answers script-valid by construction with
validity intervals no real corpus offers. The residual is external validity:
generated users are tidier than real ones, the renderer's messiness knobs
were not pushed, and results should be read as architecture comparison under
controlled conditions; a real-user study is future work.

\paragraph{LLM judge.} The judge is versioned, treated as part of the
benchmark, and held fixed across backends, so \emph{shared} error cancels
in differential claims \citep{zheng2023judging,wang2024notfair}. That does
not cover judge error that interacts with answer style --- verbosity,
abstention wording, prose vs.\ graph-derived evidence --- which is
backend-correlated, and matters most where margins are thin (the
hybrid-v2--map short-horizon interval starts at $+0.2$ pp). Judge error is
nonzero and measured only anecdotally (3 errors caught in ${\sim}200$
hand-spot-checked verdicts). As mitigation we ran a \textbf{complete
cross-family re-judging}: every answer for every headline backend, the
full-history baseline, and (in a follow-up pass) the vector backend ---
8,985 answers across both corpora, with no selection on the original
verdicts --- was independently re-judged by DeepSeek v3.2, a different
model family from answerer and judge. Overall agreement was 87.9\%
(Cohen's $\kappa=0.60$); the disagreements are strongly asymmetric ---
1{,}029 Sonnet-correct/DeepSeek-wrong vs.\ 54 the reverse --- i.e.,
DeepSeek is systematically stricter (absolute scores ${\sim}8$--12 pp
lower; per-judge aggregates and confusion counts in
Appendix~\ref{app:crossjudge}), quantifying why we treat absolute levels
as judge-defined throughout. Under DeepSeek-only verdicts:
\textbf{judge-robust} --- the graph--map crossover (week-3 endpoint
$p=0.010$, week-9 endpoint $p<10^{-5}$, interaction positive for all six
users, exact sign-flip $p=0.031$), hybrid v2 $>$ graph short-horizon
($p=0.018$), and full history tying-or-beating the memory systems at the
short horizon; \textbf{judge-sensitive} --- the hybrid-v2--map
short-horizon margin (direction preserved, $p=0.15$) and the week-9
hybrid-v2--full-history comparison (direction reverses; \S\ref{sec:cost})
--- both reported as such where they appear. A blinded \emph{human}
adjudication (pack prepared) remains the arbiter for the judge-sensitive
pair before archival submission. The corpus generator (Nova Pro) is a
different model family from the answerer (Haiku) and judge (Sonnet); the
answerer and judge share a family.

\paragraph{Benchmark-informed iteration.} Hybrid v2 was designed after
observing hybrid v1's failure and the backend results on the same
short-horizon corpus, so its 96.8\% is partly a development-set number and
is labeled as such (\S\ref{sec:layered}). Two facts limit the damage: the
six long-horizon users were generated \emph{after} hybrid v2 was fixed
(new personas and preference chains; the architecture was not re-tuned for
them), so the tenure result has a meaningful held-out character; and the
released, seeded generator makes a fully hidden test corpus cheap to
produce --- which, together with a run on an external benchmark, is the
planned pre-archival hardening.

\paragraph{Uncertainty and clustering.} Headline comparisons are reported
with exact paired McNemar tests and user-cluster bootstrap intervals
(\S\ref{sec:layered}, \S\ref{sec:crossover}); questions within a user share
facts, so users are the clustering unit, and item-level McNemar $p$-values
should be read as optimistic relative to cluster-level inference. The
tenure analysis has only six user clusters: bootstrap coverage at that size
is discrete, which is why \S\ref{sec:crossover} also reports all per-user
interaction effects and exact sign-flip randomization --- and why the
graph--map interaction sits at that test's resolution floor ($p=0.063$).
Category-level comparisons (some decided by one or two answers) are
descriptive only, and we flag the week-6 ``tie'' as a 2-answer gap rather
than an estimate of equality. Per-user and per-replicate distributions are
reported in Appendix~\ref{app:distributions}.

\paragraph{Single answerer.} All comparisons fix one answering model ---
exactly the control MemDelta shows is necessary for interpretable
comparisons \citep{wang2026memdelta}, but it scopes generality: the
write-path and topology findings are shown for one answerer. A second
answerer on the tenure subset is the highest-value robustness addition and
is planned.

\paragraph{Sample structure.} The crossover is replicated at 6 users
$\times$ 3 replicates ($n{=}324$/cell). Each user is an independent generator
draw (persona and full preference chain); the three stochastic replicates replicate
answer/judge stochasticity over the fixed corpora, not fresh content.
Chapter-level work facts, however, are derived by chapter index, so
same-archetype users share that skeleton, and the archetype draw was
concentrated (4/2/0 across three archetypes). The
inversion is robust across the variation present; per-replicate work-fact draws
and archetype balancing are the cleanest next instrument fix.

\paragraph{Injection realism.} All injection material is planted and
template-grade, in a benign harness: 14 unique probe questions (one per
user, evaluated over three stochastic trials each --- 42 answer trials)
plus the six-rung ladder --- the documented attack class
\citep{greshake2023injection,chen2024agentpoison}, not adaptive
adversaries, and not the laundering constructions of \citet{louck2026tmanm}
(one of which our episodic-leak finding independently exhibits). Zero
failures on 14 unique probes bounds the per-probe assertion rate only
below ${\sim}19\%$ (95\%, independence assumed --- optimistic, as probes
share a template family). Generating substantially more unique probes
across multiple attack templates would strengthen this result more than
further stochastic reruns of the same prompts. ``Resists the attack
class'' is the claim; ``secure'' is not. Security must also be read jointly
with clean third-party recall: the answerability stage excluded five
questions precisely because blanket third-party distrust suppresses
legitimate received information (\S 3.3) --- a system can look safe partly
by refusing; the graded-trust experiment (\S\ref{sec:security}) is the
start of quantifying that trade-off.

\paragraph{Short-horizon compression.} At ${\sim}30$--60 records/user the
corpus sits below ConvoMem's full-context threshold \citep{convomem2025},
compressing short-horizon deltas between backends --- no longer only an
inference: our full-history baseline ties the best memory system in that
regime (\S\ref{sec:cost}). This is precisely why the tenure results carry
more weight than the bake-off, and why we report both. Raw-chunk RAG and
an oracle-retrieval ceiling remain unrun.

\section{Conclusion and future work}
\label{sec:conclusion}

We inverted the benchmark pipeline --- truth before text --- and ran memory
long. The instrument's features (validity intervals, trust-typed channels,
benign injection probes, as-of-date sets) surfaced phenomena the standard
pipeline cannot see: a replicated ranking inversion with history length, an
association between write-stage errors and downstream misses aligned
fact-for-fact, injection resistance that tracked whether provenance
boundaries survive representation, and a mismatch between overall accuracy
and conditional abstention behavior. The layered architecture these findings
select is released as an open-source library with the harness as its
regression suite.

Future work, in priority order: an evidence-grounded abstention-gate
evaluation (the mechanism \S\ref{sec:security} motivates; the artifact
implements a first version); a second answerer on the tenure subset; a run
on LongMemEval's update/abstention tracks to place the design in the
field's comparison set; adaptive and laundering-grade injection
\citep[bridging to][]{louck2026tmanm,dash2026mpbench}; per-replicate work-fact
draws and archetype balancing in the generator; compaction-loss and
expiry-driven-confirmation experiments; and a real-user longitudinal study.

\section*{Reproducibility statement}

We release: the corpus generator (deterministic, seeded; MIT), the full
harness with per-cell model configurations, corpus-generator seeds, and
stochastic-replicate identifiers (MIT), the synthetic
corpus (CC BY 4.0), all judged verdicts (both judges), the complete
per-call logs (model identifiers, tags, token counts; decoding at provider
defaults), and \veracium (MIT). Exact model identifiers appear in
\S\ref{sec:harness}. The \veracium library is public at
\url{https://github.com/veracium-ai/Veracium}; the study-artifact package
(generator, harness, corpora, verdicts, logs) is in preparation at the
time of this preprint and will accompany the archival version with an
immutable release tag and an archived snapshot DOI. The paper's numbers
re-derive from the artifacts by script. The real-chat
robustness corpus (LMSYS) is license-gated and is \emph{not} redistributed;
the robustness tier ships with synthetic fixtures instead, and nothing in
the released corpus derives from LMSYS data.

\section*{Ethics and data statement}

The corpus is fully synthetic and fictionalized: no real personal data, no
scraped conversations, no real entities; injection content is
template-grade and defensive in intent (it tests quarantine, not attack
generation). Third-party-authored text is treated as untrusted \emph{by
design}, which we believe is the safety-relevant default for deployed
memory systems.

\section*{AI-collaboration disclosure}

This work was conducted as a human-directed, AI-executed collaboration: the
author set research questions, made all design decisions, and verified
results; AI assistants (Claude-family models) executed experiments, drafted
analyses, and wrote code under direction, with every paper-bound figure
re-derived from raw artifacts by script before submission.

%% file: 07_appendix.tex
\appendix

\section{Per-user and per-replicate distributions}
\label{app:distributions}

Pooled percentages compress the variation reviewers need to assess
clustering; Tables~\ref{tab:peruser} and~\ref{tab:tenureuser} report the
underlying distributions. Per-replicate short-horizon totals are stable (none
15.3/15.3/14.9\%; map 93.8/94.2/94.5\%; vector 89.5/91.6/92.0\%; graph
92.0/92.4/95.3\%; hybrid v1 87.6/88.7/92.0\%; hybrid v2
96.7/98.5/95.3\%).

\begin{table}[h]
\centering
\small
\caption{Short-horizon accuracy (\%) per user, pooled over 3 replicates
($n{=}57$--60 judged answers per cell; dov/mara/renata have hand-built
question sets, g1--g11 the generated 20-question set).}
\label{tab:peruser}
\begin{tabular}{lrrrrrr}
\toprule
User & none & map & vector & graph & hyb.\ v1 & hyb.\ v2 \\
\midrule
dov    & 15 & 83  & 88 & 82  & 80 & \textbf{98} \\
g1     & 15 & \textbf{95} & 88 & 87  & 82 & 92 \\
g2     & 16 & 93  & 91 & 93  & 91 & \textbf{95} \\
g3     & 16 & 91  & 91 & 98  & 96 & \textbf{100} \\
g4     & 15 & 92  & \textbf{95} & 92  & 88 & \textbf{95} \\
g5     & 16 & 93  & 96 & 93  & 93 & \textbf{100} \\
g6     & 16 & 98  & 96 & 96  & 89 & \textbf{100} \\
g7     & 15 & 93  & 88 & \textbf{98} & \textbf{98} & 97 \\
g8     & 16 & 95  & 93 & \textbf{100} & 93 & 98 \\
g9     & 13 & \textbf{98} & 87 & 92  & 90 & 95 \\
g10    & 15 & \textbf{100} & 88 & 93  & 92 & 97 \\
g11    & 15 & \textbf{98} & 95 & 93  & 93 & \textbf{98} \\
mara   & 15 & \textbf{100} & 90 & 98  & 93 & 98 \\
renata & 15 & 88  & 87 & 90  & 73 & \textbf{93} \\
\bottomrule
\end{tabular}
\end{table}

\begin{table}[h]
\centering
\small
\caption{Tenure accuracy (\%) per user and checkpoint, pooled over 3 replicates
($n{=}54$/cell). Cell format: map\,/\,vector\,/\,graph\,/\,hybrid v2.}
\label{tab:tenureuser}
\begin{tabular}{lccc}
\toprule
User & week 3 & week 6 & week 9 \\
\midrule
L101 & 81/70/72/78 & 76/57/83/85 & 72/69/100/98 \\
L102 & 87/80/78/87 & 83/74/80/87 & 76/76/83/83 \\
L103 & 83/78/85/83 & 85/80/80/87 & 74/76/85/87 \\
L104 & 76/70/78/83 & 69/59/81/89 & 80/61/81/93 \\
L105 & 85/81/78/76 & 85/80/78/85 & 85/81/98/100 \\
L106 & 74/72/65/74 & 80/63/72/81 & 83/85/94/98 \\
\bottomrule
\end{tabular}
\end{table}

\begin{table}[h]
\centering
\small
\caption{History-baseline tenure accuracy (\%) per user (\S\ref{sec:cost}),
pooled over 3 replicates. Cell format: full history\,/\,recency window.
Per-user week-9 hybrid v2 $-$ full history effects: $+3.7, -1.9, +7.4,
+5.6, +7.4, +3.7$ pp (L101--L106); five of six positive, exact six-user
sign-flip $p=0.063$.}
\label{tab:baselineuser}
\begin{tabular}{lccc}
\toprule
User & week 3 & week 6 & week 9 \\
\midrule
L101 & 81/56 & 87/46 & 94/69 \\
L102 & 89/48 & 93/43 & 85/65 \\
L103 & 85/52 & 87/37 & 80/65 \\
L104 & 80/39 & 81/43 & 87/65 \\
L105 & 83/50 & 89/41 & 93/67 \\
L106 & 76/52 & 78/48 & 94/69 \\
\bottomrule
\end{tabular}
\end{table}

Short-horizon per-user baselines (full history\,/\,window, \%): dov 98/80,
g1 98/73, g2 96/70, g3 98/65, g4 100/68, g5 91/68, g6 100/65, g7 98/75,
g8 100/74, g9 97/75, g10 98/62, g11 97/73, mara 98/90, renata 100/82.

\section{Cross-family judge: agreement and aggregates}
\label{app:crossjudge}

Confusion over all 8{,}985 re-judged answers --- five backends
$\times$ 1{,}797 (Sonnet $\times$ DeepSeek): both-correct 6{,}827;
Sonnet-correct/DeepSeek-wrong 1{,}029; DeepSeek-correct/Sonnet-wrong 54;
both-wrong 1{,}075. Observed agreement 87.9\%; Cohen's $\kappa=0.600$.
Agreement conditioned on the primary verdict: 86.9\% among Sonnet-correct
answers, 95.2\% among Sonnet-wrong --- the second judge rarely rescues an
answer the primary judge rejected, and its disagreements are concentrated
in added strictness. (The four-backend subset of the first pass alone:
agreement 88.1\%, $\kappa=0.574$.)

\begin{table}[h]
\centering
\small
\caption{Aggregate accuracy (\%) under each judge (same answers, both
corpora), with per-backend confusion counts (both corpora pooled,
$n{=}1{,}797$/backend): BC = both-correct, SO = Sonnet-only correct,
DO = DeepSeek-only correct, BW = both-wrong.}
\label{tab:crossjudgeagg}
\begin{tabular}{lcccccccc}
\toprule
 & \multicolumn{2}{c}{short (825)} & \multicolumn{2}{c}{tenure (972)} &
 \multicolumn{4}{c}{confusion} \\
Backend & Son. & DS & Son. & DS & BC & SO & DO & BW \\
\midrule
curated map  & 94.2 & 82.2 & 79.7 & 67.2 & 1322 & 230 &  9 & 236 \\
vector       & 91.0 & 78.8 & 72.9 & 61.8 & 1241 & 219 & 10 & 327 \\
graph        & 93.2 & 81.2 & 81.8 & 70.6 & 1353 & 211 &  3 & 230 \\
hybrid v2    & 96.8 & 84.5 & 86.4 & 72.4 & 1392 & 247 &  9 & 149 \\
full history & 97.9 & 93.0 & 85.7 & 79.7 & 1519 & 122 & 23 & 133 \\
\bottomrule
\end{tabular}
\end{table}

\paragraph{Abstention classifier.} The abstention/wrong-specific split
uses a deterministic regular expression over the answer text, applied only
to answers the judge marked wrong:
\texttt{don't have | no (information|record|mention|notes) | don't know |
not in (my|the) notes | no such | unable to} (case-insensitive). An answer
matching anywhere is classified as an abstention, otherwise as a wrong
specific --- so a hedged answer that both abstains and asserts counts as
an abstention, and partial answers are classified by the presence of any
disclaimer phrase, a known coarseness. The classifier is reproducible but
has not been validated against human labels; the planned blinded
adjudication uses a three-way rubric (correct / wrong-specific /
abstention) partly to measure this classifier's agreement with human
judgment.

Confabulation metrics under both judges (tenure; wrong-set defined by each
judge): conditional confident-error share --- vector 62.7\% (Sonnet) /
62.5\% (DeepSeek); map 72.1 / 69.3; graph 83.6 / 78.0; hybrid v2 93.9 /
85.4. Unconditional wrong-specific rate --- vector 17.0 / 23.9; map 14.6 /
22.7; graph 15.2 / 22.9; hybrid v2 12.8 / 23.6; full history (DeepSeek)
18.1. The conditional ordering (hybrid v2 most confident when wrong,
vector most abstaining) is stable across judges; the unconditional
ordering is judge-sensitive.

%% file: main.bbl
\begin{thebibliography}{56}
\providecommand{\natexlab}[1]{#1}
\providecommand{\url}[1]{\texttt{#1}}
\expandafter\ifx\csname urlstyle\endcsname\relax
  \providecommand{\doi}[1]{doi: #1}\else
  \providecommand{\doi}{doi: \begingroup \urlstyle{rm}\Url}\fi

\bibitem[Bhargava and Sobral~Barrento(2026)]{bhargava2026memaudit}
Nishant Bhargava and Rodrigo Sobral~Barrento.
\newblock {MEMAUDIT}: An exact package-oracle evaluation protocol for budgeted
  long-term {LLM} memory writing.
\newblock \emph{arXiv preprint arXiv:2605.02199}, 2026.
\newblock URL \url{https://arxiv.org/abs/2605.02199}.

\bibitem[Chen et~al.(2024)Chen, Xiang, Xiao, Song, and Li]{chen2024agentpoison}
Zhaorun Chen, Zhen Xiang, Chaowei Xiao, Dawn Song, and Bo~Li.
\newblock {AgentPoison}: Red-teaming {LLM} agents via poisoning memory or
  knowledge bases.
\newblock In \emph{Advances in Neural Information Processing Systems
  (NeurIPS)}, 2024.
\newblock URL \url{https://arxiv.org/abs/2407.12784}.

\bibitem[Chhikara et~al.(2025)Chhikara, Khant, Aryan, Singh, and
  Yadav]{chhikara2025mem0}
Prateek Chhikara, Dev Khant, Saket Aryan, Taranjeet Singh, and Deshraj Yadav.
\newblock {Mem0}: Building production-ready {AI} agents with scalable long-term
  memory.
\newblock \emph{arXiv preprint arXiv:2504.19413}, 2025.
\newblock URL \url{https://arxiv.org/abs/2504.19413}.

\bibitem[Dash et~al.(2026)Dash, Ge, Jain, Shah, and Shang]{dash2026mpbench}
Pritam Dash, Tongyu Ge, Aditi Jain, Tanmay Shah, and Zhiwei Shang.
\newblock From untrusted input to trusted memory: A systematic study of memory
  poisoning attacks in {LLM} agents.
\newblock \emph{arXiv preprint arXiv:2606.04329}, 2026.
\newblock URL \url{https://arxiv.org/abs/2606.04329}.

\bibitem[Deng et~al.(2026)Deng, Zhong, Peng, Lu, Wu, Li, Xu, Yao, Fang, Cao,
  Guo, Yuan, Ma, Yu, Hu, Dong, Zhu, and Zhang]{deng2026memtrace}
Xinle Deng, Ruobin Zhong, Hujin Peng, Xiaoben Lu, Yanzhe Wu, Guang Li, Buqiang
  Xu, Yunzhi Yao, Jizhan Fang, Haoliang Cao, Junjie Guo, Yuan Yuan, Ziqing Ma,
  Yuanqiang Yu, Rui Hu, Baohua Dong, Hangcheng Zhu, and Ningyu Zhang.
\newblock {MemTrace}: Tracing and attributing errors in large language model
  memory systems.
\newblock \emph{arXiv preprint arXiv:2605.28732}, 2026.
\newblock URL \url{https://arxiv.org/abs/2605.28732}.

\bibitem[Derehag et~al.(2026)Derehag, Calva, and
  Ghiurau]{derehag2026smartsearch}
Jesper Derehag, Carlos Calva, and Timmy Ghiurau.
\newblock {SmartSearch}: How ranking beats structure for conversational memory
  retrieval.
\newblock \emph{arXiv preprint arXiv:2603.15599}, 2026.
\newblock URL \url{https://arxiv.org/abs/2603.15599}.

\bibitem[Farquhar et~al.(2024)Farquhar, Kossen, Kuhn, and
  Gal]{farquhar2024semantic}
Sebastian Farquhar, Jannik Kossen, Lorenz Kuhn, and Yarin Gal.
\newblock Detecting hallucinations in large language models using semantic
  entropy.
\newblock \emph{Nature}, 630\penalty0 (8017):\penalty0 625--630, 2024.
\newblock URL \url{https://www.nature.com/articles/s41586-024-07421-0}.

\bibitem[Flynt(2026)]{flynt2026orgforge}
Jeffrey Flynt.
\newblock {OrgForge-IT}: A verifiable synthetic benchmark for {LLM}-based
  insider threat detection.
\newblock \emph{arXiv preprint arXiv:2603.22499}, 2026.
\newblock URL \url{https://arxiv.org/abs/2603.22499}.

\bibitem[Greshake et~al.(2023)Greshake, Abdelnabi, Mishra, Endres, Holz, and
  Fritz]{greshake2023injection}
Kai Greshake, Sahar Abdelnabi, Shailesh Mishra, Christoph Endres, Thorsten
  Holz, and Mario Fritz.
\newblock Not what you've signed up for: Compromising real-world
  {LLM}-integrated applications with indirect prompt injection.
\newblock In \emph{Proceedings of the 16th ACM Workshop on Artificial
  Intelligence and Security (AISec)}, 2023.
\newblock URL \url{https://arxiv.org/abs/2302.12173}.

\bibitem[Gu et~al.(2026)Gu, Zhang, Khattab, and Madden]{gu2026peek}
Zhuohan Gu, Qizheng Zhang, Omar Khattab, and Samuel Madden.
\newblock {PEEK}: Context map as an orientation cache for long-context {LLM}
  agents.
\newblock \emph{arXiv preprint arXiv:2605.19932}, 2026.
\newblock URL \url{https://arxiv.org/abs/2605.19932}.

\bibitem[Guti\'{e}rrez et~al.(2024)Guti\'{e}rrez, Shu, Gu, Yasunaga, and
  Su]{gutierrez2024hipporag}
Bernal~Jim\'{e}nez Guti\'{e}rrez, Yiheng Shu, Yu~Gu, Michihiro Yasunaga, and
  Yu~Su.
\newblock {HippoRAG}: Neurobiologically inspired long-term memory for large
  language models.
\newblock In \emph{Advances in Neural Information Processing Systems
  (NeurIPS)}, 2024.
\newblock URL \url{https://arxiv.org/abs/2405.14831}.

\bibitem[He et~al.(2026)He, Wang, Zhi, Hu, Chen, Yin, Chen, Wu, Ouyang, Wang,
  Pei, McAuley, Choi, and Pentland]{he2026memoryarena}
Zexue He, Yu~Wang, Churan Zhi, Yuanzhe Hu, Tzu-Ping Chen, Lang Yin, Ze~Chen,
  Tong~Arthur Wu, Siru Ouyang, Zihan Wang, Jiaxin Pei, Julian McAuley, Yejin
  Choi, and Alex Pentland.
\newblock {MemoryArena}: Benchmarking agent memory in interdependent
  multi-session agentic tasks.
\newblock \emph{arXiv preprint arXiv:2602.16313}, 2026.
\newblock URL \url{https://arxiv.org/abs/2602.16313}.

\bibitem[Hu et~al.(2025)Hu, Wang, and McAuley]{hu2026memoryagentbench}
Yuanzhe Hu, Yu~Wang, and Julian McAuley.
\newblock Evaluating memory in {LLM} agents via incremental multi-turn
  interactions.
\newblock \emph{arXiv preprint arXiv:2507.05257}, 2025.
\newblock URL \url{https://arxiv.org/abs/2507.05257}.
\newblock MemoryAgentBench; ICLR 2026.

\bibitem[Jiang et~al.(2026)Jiang, Li, Wei, Yang, Kishore, Zhao, Kang, Hu, Chen,
  Li, and Li]{jiang2026anatomy}
Dongming Jiang, Yi~Li, Songtao Wei, Jinxin Yang, Ayushi Kishore, Alysa Zhao,
  Dingyi Kang, Xu~Hu, Feng Chen, Qiannan Li, and Bingzhe Li.
\newblock Anatomy of agentic memory: Taxonomy and empirical analysis of
  evaluation and system limitations.
\newblock \emph{arXiv preprint arXiv:2602.19320}, 2026.
\newblock URL \url{https://arxiv.org/abs/2602.19320}.

\bibitem[Kadavath et~al.(2022)Kadavath, Conerly, Askell,
  et~al.]{kadavath2022know}
Saurav Kadavath, Tom Conerly, Amanda Askell, et~al.
\newblock Language models (mostly) know what they know.
\newblock \emph{arXiv preprint arXiv:2207.05221}, 2022.
\newblock URL \url{https://arxiv.org/abs/2207.05221}.

\bibitem[Lam et~al.(2026)Lam, Li, Zhang, and Zhao]{lam2026ssgm}
Chingkwun Lam, Jiaxin Li, Lingfei Zhang, and Kuo Zhao.
\newblock Governing evolving memory in {LLM} agents: Risks, mechanisms, and the
  stability and safety governed memory ({SSGM}) framework.
\newblock \emph{arXiv preprint arXiv:2603.11768}, 2026.
\newblock URL \url{https://arxiv.org/abs/2603.11768}.

\bibitem[Li et~al.(2026)Li, Li, Ma, Zhu, Liu, Wang, and Feng]{li2026plant}
Yongxiang Li, Moxin Li, Zhixin Ma, Fengbin Zhu, Dongrui Liu, Wenjie Wang, and
  Fuli Feng.
\newblock Plant, persist, trigger: Sleeper attack on large language model
  agents.
\newblock \emph{arXiv preprint arXiv:2605.28201}, 2026.
\newblock URL \url{https://arxiv.org/abs/2605.28201}.

\bibitem[Lin et~al.(2026)Lin, Hao, Fu, Cui, Chen, Li, Li, and
  Xiong]{lin2026ltmsecurity}
Zehao Lin, Xixuan Hao, Renyu Fu, Shaobo Cui, Kai Chen, Chunyu Li, Zhiyu Li, and
  Feiyu Xiong.
\newblock A survey on long-term memory security in {LLM} agents: Attacks,
  defenses, and governance across the memory lifecycle.
\newblock \emph{arXiv preprint arXiv:2604.16548}, 2026.
\newblock URL \url{https://arxiv.org/abs/2604.16548}.

\bibitem[Louck(2026)]{louck2026tmanm}
Yedidel Louck.
\newblock Securing {LLM}-agent long-term memory against poisoning:
  Non-malleable, origin-bound authority with machine-checked guarantees.
\newblock \emph{arXiv preprint arXiv:2606.24322}, 2026.
\newblock URL \url{https://arxiv.org/abs/2606.24322}.

\bibitem[Lu et~al.(2026)Lu, Li, Shi, Wang, Wang, and Hu]{lu2026seem}
Zhengxuan Lu, Dongfang Li, Yukun Shi, Beilun Wang, Longyue Wang, and Baotian
  Hu.
\newblock Structured episodic event memory.
\newblock \emph{arXiv preprint arXiv:2601.06411}, 2026.
\newblock URL \url{https://arxiv.org/abs/2601.06411}.

\bibitem[Maharana et~al.(2024)Maharana, Lee, Tulyakov, Bansal, Barbieri, and
  Fung]{maharana2024locomo}
Adyasha Maharana, Dong-Ho Lee, Sergey Tulyakov, Mohit Bansal, Francesco
  Barbieri, and Yuwei Fung.
\newblock Evaluating very long-term conversational memory of {LLM} agents.
\newblock In \emph{Proceedings of the 62nd Annual Meeting of the Association
  for Computational Linguistics (ACL)}, 2024.
\newblock URL \url{https://arxiv.org/abs/2402.17753}.

\bibitem[{Mem0}(2026)]{mem0report2026}
{Mem0}.
\newblock {AI} memory benchmarks in 2026: {LoCoMo}, {LongMemEval} \& {BEAM}.
\newblock Industry report (blog), 2026.
\newblock URL \url{https://mem0.ai/blog/ai-memory-benchmarks-in-2026}.

\bibitem[Meng et~al.(2022)Meng, Bau, Andonian, and Belinkov]{meng2022rome}
Kevin Meng, David Bau, Alex Andonian, and Yonatan Belinkov.
\newblock Locating and editing factual associations in {GPT}.
\newblock In \emph{Advances in Neural Information Processing Systems
  (NeurIPS)}, 2022.
\newblock URL \url{https://arxiv.org/abs/2202.05262}.

\bibitem[Meng et~al.(2023)Meng, Sen~Sharma, Andonian, Belinkov, and
  Bau]{meng2023memit}
Kevin Meng, Arnab Sen~Sharma, Alex Andonian, Yonatan Belinkov, and David Bau.
\newblock Mass-editing memory in a transformer.
\newblock In \emph{International Conference on Learning Representations
  (ICLR)}, 2023.
\newblock URL \url{https://arxiv.org/abs/2210.07229}.

\bibitem[Northcutt et~al.(2021)Northcutt, Athalye, and
  Mueller]{northcutt2021labelerrors}
Curtis~G. Northcutt, Anish Athalye, and Jonas Mueller.
\newblock Pervasive label errors in test sets destabilize machine learning
  benchmarks.
\newblock In \emph{NeurIPS Datasets and Benchmarks Track}, 2021.
\newblock URL \url{https://arxiv.org/abs/2103.14749}.

\bibitem[{OWASP GenAI Security Project}(2025)]{owasp2025llm01}
{OWASP GenAI Security Project}.
\newblock {LLM01}:2025 prompt injection, 2025.
\newblock URL \url{https://genai.owasp.org/llmrisk/llm01-prompt-injection/}.

\bibitem[Packer et~al.(2023)Packer, Wooders, Lin, Fang, Patil, Stoica, and
  Gonzalez]{packer2023memgpt}
Charles Packer, Sarah Wooders, Kevin Lin, Vivian Fang, Shishir~G. Patil, Ion
  Stoica, and Joseph~E. Gonzalez.
\newblock {MemGPT}: Towards {LLMs} as operating systems.
\newblock \emph{arXiv preprint arXiv:2310.08560}, 2023.
\newblock URL \url{https://arxiv.org/abs/2310.08560}.

\bibitem[Pakhomov et~al.(2025)Pakhomov, Nijkamp, and Xiong]{convomem2025}
Egor Pakhomov, Erik Nijkamp, and Caiming Xiong.
\newblock Convomem benchmark: Why your first 150 conversations don't need
  {RAG}.
\newblock \emph{arXiv preprint arXiv:2511.10523}, 2025.
\newblock URL \url{https://arxiv.org/abs/2511.10523}.

\bibitem[Park et~al.(2023)Park, O'Brien, Cai, Morris, Liang, and
  Bernstein]{park2023generative}
Joon~Sung Park, Joseph~C. O'Brien, Carrie~J. Cai, Meredith~Ringel Morris, Percy
  Liang, and Michael~S. Bernstein.
\newblock Generative agents: Interactive simulacra of human behavior.
\newblock In \emph{Proceedings of the 36th Annual ACM Symposium on User
  Interface Software and Technology (UIST)}, 2023.
\newblock URL \url{https://arxiv.org/abs/2304.03442}.

\bibitem[Patel(2026)]{patel2026supersede}
Vedant Patel.
\newblock Supersede: Diagnosing and training the memory-update gap in {LLM}
  agents.
\newblock \emph{arXiv preprint arXiv:2606.27472}, 2026.
\newblock URL \url{https://arxiv.org/abs/2606.27472}.

\bibitem[{Penfield Labs}(2026)]{penfield2026locomoaudit}
{Penfield Labs}.
\newblock We audited {LoCoMo}: 6.4\% of the answer key is wrong and the judge
  accepts up to 63\%.
\newblock Blog post; reproducible audit: github.com/dial481/locomo-audit, 2026.
\newblock URL
  \url{https://penfieldlabs.substack.com/p/we-audited-locomo-64-of-the-answer}.

\bibitem[{Perplexity AI}(2026)]{perplexity2026brain}
{Perplexity AI}.
\newblock Self-improving memory for agents (``brain'').
\newblock Perplexity blog, 2026.
\newblock URL
  \url{https://www.perplexity.ai/hub/blog/self-improving-memory-for-agents}.

\bibitem[Pulipaka et~al.(2026)Pulipaka, Hlebik, Raghav, Abdelnabi, Raina,
  Sheth, and Fritz]{pulipaka2026sleeper}
Sidharth Pulipaka, Stanislau Hlebik, Leonidas Raghav, Sahar Abdelnabi, Vyas
  Raina, Ivaxi Sheth, and Mario Fritz.
\newblock Hidden in memory: Sleeper memory poisoning in {LLM} agents.
\newblock \emph{arXiv preprint arXiv:2605.15338}, 2026.
\newblock URL \url{https://arxiv.org/abs/2605.15338}.

\bibitem[Rasmussen et~al.(2025)Rasmussen, Paliychuk, Beauvais, Ryan, and
  Chalef]{rasmussen2025zep}
Preston Rasmussen, Pavlo Paliychuk, Travis Beauvais, Jack Ryan, and Daniel
  Chalef.
\newblock {Zep}: A temporal knowledge graph architecture for agent memory.
\newblock \emph{arXiv preprint arXiv:2501.13956}, 2025.
\newblock URL \url{https://arxiv.org/abs/2501.13956}.

\bibitem[Reddy and Challaram(2026)]{reddy2026freshness}
Vikas Reddy and Sumanth Challaram.
\newblock Don't ask the {LLM} to track freshness: A deterministic recipe for
  memory conflict resolution.
\newblock \emph{arXiv preprint arXiv:2606.01435}, 2026.
\newblock URL \url{https://arxiv.org/abs/2606.01435}.

\bibitem[Roig(2026)]{roig2026riker}
JV~Roig.
\newblock Scalable and reliable evaluation of {AI} knowledge retrieval systems:
  {RIKER} and the coherent simulated universe.
\newblock \emph{arXiv preprint arXiv:2601.08847}, 2026.
\newblock URL \url{https://arxiv.org/abs/2601.08847}.

\bibitem[Roynard(2026)]{roynard2026knowledgelayer}
Micha\"{e}l Roynard.
\newblock The missing knowledge layer in cognitive architectures for ai agents.
\newblock \emph{arXiv preprint arXiv:2604.11364}, 2026.
\newblock URL \url{https://arxiv.org/abs/2604.11364}.

\bibitem[Shen et~al.(2026)Shen, Li, Zhou, and Hu]{shen2026mem2actbench}
Yiting Shen, Kun Li, Wei Zhou, and Songlin Hu.
\newblock {Mem2ActBench}: A benchmark for evaluating long-term memory
  utilization in task-oriented autonomous agents.
\newblock \emph{arXiv preprint arXiv:2601.19935}, 2026.
\newblock URL \url{https://arxiv.org/abs/2601.19935}.

\bibitem[Shutova et~al.(2026)Shutova, Olenina, Vinogradov, and
  Sinitsin]{shutova2026structmemeval}
Alina Shutova, Alexandra Olenina, Ivan Vinogradov, and Anton Sinitsin.
\newblock Evaluating memory structure in {LLM} agents.
\newblock \emph{arXiv preprint arXiv:2602.11243}, 2026.
\newblock URL \url{https://arxiv.org/abs/2602.11243}.

\bibitem[Sun et~al.(2026)Sun, Zhu, Yao, Liu, and Han]{trimem2026}
Jingwei Sun, Jianing Zhu, Jiangchao Yao, Tongliang Liu, and Bo~Han.
\newblock {TriMem}: Rethinking how to remember---beyond atomic facts in
  lifelong {LLM} agent memory.
\newblock \emph{arXiv preprint arXiv:2605.19952}, 2026.
\newblock URL \url{https://arxiv.org/abs/2605.19952}.

\bibitem[Tan et~al.(2026)Tan, Yao, Jin, Yu, Wang, Fan, Lu, Liu, Zhang, Ma,
  Yang, and Sun]{tan2026memaudit}
Zhewen Tan, Yilun Yao, Huiyan Jin, Wenhan Yu, Guoan Wang, Mengyuan Fan, Liang
  Lu, Feng Liu, Xiangzheng Zhang, Duohe Ma, Tong Yang, and Lin Sun.
\newblock {MemAudit}: Post-hoc auditing of poisoned agent memory via causal
  attribution and structural anomaly detection.
\newblock \emph{arXiv preprint arXiv:2605.23723}, 2026.
\newblock URL \url{https://arxiv.org/abs/2605.23723}.

\bibitem[Tavakoli et~al.(2025)Tavakoli, Salemi, Ye, Abdalla, Zamani, and
  Mitchell]{tavakoli2025beam}
Mohammad Tavakoli, Alireza Salemi, Carrie Ye, Mohamed Abdalla, Hamed Zamani,
  and J~Ross Mitchell.
\newblock Beyond a million tokens: Benchmarking and enhancing long-term memory
  in {LLMs}.
\newblock \emph{arXiv preprint arXiv:2510.27246}, 2025.
\newblock URL \url{https://arxiv.org/abs/2510.27246}.

\bibitem[Wang(2026)]{wang2026memdelta}
Kuan Wang.
\newblock {MemDelta}: Controlled baselines and hidden confounds in agent memory
  evaluation.
\newblock \emph{arXiv preprint arXiv:2606.29914}, 2026.
\newblock URL \url{https://arxiv.org/abs/2606.29914}.

\bibitem[Wang et~al.(2024)Wang, Li, Chen, Cai, Zhu, Lin, Cao, Liu, Liu, and
  Sui]{wang2024notfair}
Peiyi Wang, Lei Li, Liang Chen, Zefan Cai, Dawei Zhu, Binghuai Lin, Yunbo Cao,
  Qi~Liu, Tianyu Liu, and Zhifang Sui.
\newblock Large language models are not fair evaluators.
\newblock In \emph{Proceedings of the 62nd Annual Meeting of the Association
  for Computational Linguistics (ACL)}, 2024.
\newblock URL \url{https://arxiv.org/abs/2305.17926}.

\bibitem[Wen et~al.(2024)Wen, Yao, Feng, Xu, Tsvetkov, Howe, and
  Wang]{wen2024abstention}
Bingbing Wen, Jihan Yao, Shangbin Feng, Chenjun Xu, Yulia Tsvetkov, Bill Howe,
  and Lucy~Lu Wang.
\newblock Know your limits: A survey of abstention in large language models.
\newblock \emph{arXiv preprint arXiv:2407.18418}, 2024.
\newblock URL \url{https://arxiv.org/abs/2407.18418}.

\bibitem[Wu et~al.(2025)Wu, Wang, Yu, Zhang, Chang, and Yu]{wu2024longmemeval}
Di~Wu, Hongwei Wang, Wenhao Yu, Yuwei Zhang, Kai-Wei Chang, and Dong Yu.
\newblock {LongMemEval}: Benchmarking chat assistants on long-term interactive
  memory.
\newblock In \emph{International Conference on Learning Representations
  (ICLR)}, 2025.
\newblock URL \url{https://arxiv.org/abs/2410.10813}.

\bibitem[Xu et~al.(2024)Xu, Guan, Greene, and Kechadi]{xu2024contamination}
Cheng Xu, Shuhao Guan, Derek Greene, and M-Tahar Kechadi.
\newblock Benchmark data contamination of large language models: A survey.
\newblock \emph{arXiv preprint arXiv:2406.04244}, 2024.
\newblock URL \url{https://arxiv.org/abs/2406.04244}.

\bibitem[Xu et~al.(2025)Xu, Liang, Mei, Gao, Tan, and Zhang]{xu2025amem}
Wujiang Xu, Zujie Liang, Kai Mei, Hang Gao, Juntao Tan, and Yongfeng Zhang.
\newblock {A-Mem}: Agentic memory for {LLM} agents.
\newblock \emph{arXiv preprint arXiv:2502.12110}, 2025.
\newblock URL \url{https://arxiv.org/abs/2502.12110}.

\bibitem[Xu et~al.(2026)Xu, Wang, Mei, Liang, Wang, Jin, Zhang, Zhang, Hua,
  Sahu, and Metaxas]{xu2026memgym}
Wujiang Xu, Yu~Wang, Kai Mei, Kaiqu Liang, Zhenting Wang, Mingyu Jin, Han
  Zhang, Shi-Xiong Zhang, Wenyue Hua, Sambit Sahu, and Dimitris~N. Metaxas.
\newblock {MemGym}: a long-horizon memory environment for {LLM} agents.
\newblock \emph{arXiv preprint arXiv:2605.20833}, 2026.
\newblock URL \url{https://arxiv.org/abs/2605.20833}.

\bibitem[Yadav(2026)]{yadav2026memstrata}
Neeraj Yadav.
\newblock Temporal validity in retrieval memory: Eliminating stale-fact errors
  for {AI} agents over evolving knowledge.
\newblock \emph{arXiv preprint arXiv:2606.26511}, 2026.
\newblock URL \url{https://arxiv.org/abs/2606.26511}.

\bibitem[Yuan et~al.(2026)Yuan, Su, and Yao]{yuan2026bottlenecks}
Boqin Yuan, Yue Su, and Kun Yao.
\newblock Diagnosing retrieval vs. utilization bottlenecks in {LLM} agent
  memory.
\newblock \emph{arXiv preprint arXiv:2603.02473}, 2026.
\newblock URL \url{https://arxiv.org/abs/2603.02473}.

\bibitem[Zheng et~al.(2023)Zheng, Chiang, Sheng, Zhuang, Wu, Zhuang, Lin, Li,
  Li, Xing, Zhang, Gonzalez, and Stoica]{zheng2023judging}
Lianmin Zheng, Wei-Lin Chiang, Ying Sheng, Siyuan Zhuang, Zhanghao Wu, Yonghao
  Zhuang, Zi~Lin, Zhuohan Li, Dacheng Li, Eric~P. Xing, Hao Zhang, Joseph~E.
  Gonzalez, and Ion Stoica.
\newblock Judging {LLM}-as-a-judge with {MT-Bench} and chatbot arena.
\newblock In \emph{Advances in Neural Information Processing Systems
  (NeurIPS)}, 2023.
\newblock URL \url{https://arxiv.org/abs/2306.05685}.

\bibitem[Zhong et~al.(2024)Zhong, Guo, Gao, Ye, and Wang]{zhong2023memorybank}
Wanjun Zhong, Lianghong Guo, Qiqi Gao, He~Ye, and Yanlin Wang.
\newblock {MemoryBank}: Enhancing large language models with long-term memory.
\newblock \emph{Proceedings of the AAAI Conference on Artificial Intelligence},
  2024.
\newblock URL \url{https://arxiv.org/abs/2305.10250}.

\bibitem[Zhong et~al.(2023)Zhong, Wu, Manning, Potts, and
  Chen]{zhong2023mquake}
Zexuan Zhong, Zhengxuan Wu, Christopher~D. Manning, Christopher Potts, and
  Danqi Chen.
\newblock {MQuAKE}: Assessing knowledge editing in language models via
  multi-hop questions.
\newblock In \emph{Proceedings of the 2023 Conference on Empirical Methods in
  Natural Language Processing (EMNLP)}, 2023.
\newblock URL \url{https://arxiv.org/abs/2305.14795}.

\bibitem[Zhou et~al.(2026)Zhou, Zhou, Han, Xu, Li, Li, Xiong, and
  Wu]{zhou2026agentnative}
Wei Zhou, Xuanhe Zhou, Shaokun Han, Hongming Xu, Guoliang Li, Zhiyu Li, Feiyu
  Xiong, and Fan Wu.
\newblock Are we ready for an agent-native memory system?
\newblock \emph{arXiv preprint arXiv:2606.24775}, 2026.
\newblock URL \url{https://arxiv.org/abs/2606.24775}.

\bibitem[Zhu et~al.(2026)Zhu, Chen, Yu, Wu, and Wang]{zhu2026tiermem}
Qiming Zhu, Shunian Chen, Rui Yu, Zhehao Wu, and Benyou Wang.
\newblock From lossy to verified: A provenance-aware tiered memory for agents.
\newblock \emph{arXiv preprint arXiv:2602.17913}, 2026.
\newblock URL \url{https://arxiv.org/abs/2602.17913}.

\end{thebibliography}
